\documentclass[runningheads]{llncs}
\usepackage{graphicx}
\usepackage{float}
\usepackage{subcaption}

\newcommand{\diff}{\,\text{d}}

\newcommand\restr[2]{{
		\left.\kern-\nulldelimiterspace 
		#1 
		\vphantom{\big|} 
		\right|_{#2} 
}}

\newcommand{\tnorm}{\mathcal{T}}
\newcommand{\imp}{\mathcal{I}}
\newcommand{\pwset}{\mathcal{P}}
\newcommand{\fpwset}{\widetilde{\mathcal{P}}}

\usepackage[utf8]{inputenc}
\usepackage{lmodern}
\usepackage{mathtools}
\usepackage{amssymb}
\usepackage{esint}
\usepackage{hyperref}

\begin{document}
\title{Fuzzy Rough Choquet Distances for Classification}
%
%
\author{Adnan Theerens\orcidID{0000-0001-5412-674X} \and
	Chris Cornelis\orcidID{0000-0002-6852-4041}}
\authorrunning{A. Theerens and C. Cornelis}
%
\institute{Computational Web Intelligence, Department of Applied Mathematics, Computer Science and Statistics, Ghent University, Ghent, Belgium
	\email{\{adnan.theerens,chris.cornelis\}@ugent.be}}
\maketitle              
\begin{abstract}
	This paper introduces a novel Choquet distance using fuzzy rough set based measures.
	The proposed distance measure combines the attribute information received from fuzzy rough set theory with the flexibility of the Choquet integral. This approach is designed to adeptly capture non-linear relationships within the data, acknowledging the interplay of the conditional attributes towards the decision attribute and resulting in a more flexible and accurate distance. We explore its application in the context of machine learning, with a specific emphasis on distance-based classification approaches (e.g.\ k-nearest neighbours). The paper examines two fuzzy rough set based measures that are based on the positive region. Moreover, we explore two procedures for monotonizing the measures derived from fuzzy rough set theory, making them suitable for use with the Choquet integral, and investigate their differences.
\end{abstract}

\section{Introduction}
In the field of machine learning, there is an ongoing search for algorithms that are more effective, robust, and explainable. Many important machine learning techniques rely on the idea of measuring similarity or dissimilarity between instances. This is commonly done using distance measures, forming the basis for fundamental instance-based classification algorithms like k-nearest neighbours (KNN) and fuzzy-rough nearest neighbours (FRNN, 
\cite{jensen2011fuzzy,lenz2019scalable}), as well as algorithms such as k-means clustering and density-based spatial clustering (DBSCAN) \cite{dbscan} for unsupervised learning, and local outlier factor (LOF) \cite{breunig2000lof} for anomaly detection.

Classical distance measures, like the Euclidean or Manhattan distance, have been fundamental in traditional machine learning approaches. However, with the growing complexity and diversity of datasets, the shortcomings of these conventional measures are becoming more evident. The need for a more versatile, adaptive, and information-rich distance metric has become more apparent in order to better capture the nuanced relationships within the data. In classification tasks, it is important to ensure that the nearest neighbours belong to the same class. This is because the classification goal is to effectively separate data into distinct classes. Therefore, our aim is to bring instances of the same class closer together while pushing instances of different classes farther apart. This paper introduces a novel distance measure for classification based on the Choquet integral that does exactly this. The Choquet integral, which is a generalization of the Lebesgue integral to non-additive measures and is commonly used in decision-making \cite{grabisch2010decade}, is known for capturing non-linear relationships. In the context of supervised learning problems, it allows us to consider the interplay between the conditional attributes towards the decision attribute, providing an intuitive approach for improving supervised learning outcomes.

Choquet integrals have been used in the past to create distances. In \cite{bolton2008discrete} the authors characterize the class of measures that induce a metric through the Choquet integral. In \cite{torra2012comparison}, the Choquet integral is generalized to incorporate the Mahalanobis distance into the resulting distances. In \cite{ma2017choquetdistancesclassification} the authors proposed a nonlinear classifier based on the Choquet integral with respect to a signed efficiency measure, and the boundary is a Choquet broken-hyperplane. Moreover, \cite{abril2012choquet} introduces a novel distance based record linkage using the Choquet integral to compute the distances between records. They employ a learning approach to determine the optimal fuzzy measure for the linkage process. However, none of these studies harnesses the potential of the Choquet integral for defining distances to be used with distance-based classification algorithms. Another distinction is that previous works often approach these distances abstractly, or rely on human input for obtaining the measures, or they employ them in different applications, such as record linkage. Another notable distinction lies in the fact that the previously defined Choquet distances are based on the Euclidean distance, whereas our approach is based on the more general Minkowski distance. This allows for greater flexibility, and the inclusion of the Manhattan and Chebyshev distance, which can offer advantages in supervised learning scenarios (cf.\ \cite{lenz2023unified}).

One of the challenges of the Choquet integral is providing it with a suitable/adequate monotone measure. In the case of constructing a Choquet distance measure for supervised learning this would result in \(2^n-2\) parameters (one parameter for each subset of attributes, except the empty set and the universe), where \(n\) is the number of conditional attributes. In order to avoid a complicated learning procedure to determine the optimal measure, we will make use of fuzzy rough set based measures to evaluate how well a subset of conditional attributes relates to the decision attribute.

Fuzzy rough set theory, a combination of fuzzy set theory \cite{fuzzysets} and rough set theory \cite{pawlak1982rough}, provides a practical and effective formal framework for describing and leveraging data dependencies \cite{cornelis2010attribute,roughsetTheoryMeasure}. By incorporating fuzzy set theory's ability to handle uncertainty and rough set theory's approach to handling inconsistency, this approach proves to be a versatile tool for real-world data analysis. In particular, the ability to describe data dependenci`es on the conditional attributes makes it a logical choice for constructing monotone measures that describe the dependency of the decision attribute and a subset of conditional attributes. In the context of rough sets, such measures have already been proposed, as in \cite{roughsetTheoryMeasure} for multicriteria decision analysis. Additionally, analogous measures have been developed for fuzzy rough sets to be used for attribute selection algorithms \cite{cornelis2010attribute,cornelis2008feature}. These measures can then be used with the Choquet integral to generate distances for classification purposes.

The remainder of this paper is organized as follows: in Section \ref{sec: prelims}, we recall the required prerequisites for fuzzy rough sets and Choquet integration. Section \ref{sec: FRdistances} introduces fuzzy rough Choquet \(p\)-distances, explores different ways to define monotone measures based on fuzzy rough sets and shows how these can be used for classification.
Section \ref{sec: conclusion} concludes this paper and provides insights into further research directions.
\section{Preliminaries}
\label{sec: prelims}
\subsection{Fuzzy rough set theory}
\label{sec: fuzzy rough set theory}
We will make use of the following fuzzy logic connectives.
\begin{definition}
	\begin{itemize}
		\item A function \(\mathcal{C}:[0,1]^2\to [0,1]\) is called a \emph{conjunctor} if it is increasing in both arguments and satisfies \(\mathcal{C}(0,0)=\mathcal{C}(1,0)=\mathcal{C}(0,1)=0\), \(\mathcal{C}(1,1)=1\) and \(\mathcal{C}(1,x)=x\) for all \(x\in[0,1]\). A commutative and associative conjunctor \(\tnorm\) is called a \emph{t-norm}.
		\item A function $\imp: \left[0,1\right]^2\rightarrow \left[0,1\right]$ is called an \emph{implicator} if $\imp(0,0)=\imp(0,1)=\imp(1,1)=1$, \(\imp(1,0)=0\) and for all $x_1,x_2,y_1,y_2$ in $ \left[0,1\right]$ the following holds:
		      \begin{enumerate}
			      \item $x_1\leq x_2\Rightarrow \imp(x_1,y_1)\geq \imp(x_2,y_1)$ (decreasing in the first argument),
			      \item $y_1\leq y_2\Rightarrow \imp(x_1,y_1)\leq \imp(x_1,y_2)$ (increasing in the second argument),
		      \end{enumerate}
		\item A function \(\mathcal{N}:[0,1]\to [0,1]\) is called a \emph{negator} if it is non-increasing and satisfies \(\mathcal{N}(0)=1\) and \(\mathcal{N}(1)=0\). An implicator \(\imp\) induces a negator \(\mathcal{N}^\imp(x):= \imp(x,0)\).
	\end{itemize}
\end{definition}
Since t-norms are required to be associative, they can be extended naturally to a function \([0,1]^n\to [0,1]\) for any natural number \(n\geq 2\).

Rough sets, introduced by Pawlak \cite{pawlak1982rough}, aim to handle uncertainty related to \emph{indiscernibility}. Fuzzy rough sets extend this approach to incorporate similarity and fuzzy concepts. We will use the notation \(\pwset(X)\) to represent the powerset of \(X\) and assume \(X\) to be finite throughout this paper. Likewise, we will use the notation \(\fpwset(X)\) to represent the set consisting of all fuzzy sets on \(X\) (i.e.\ functions of the form \(X\to [0,1]\)).
\begin{definition}[Fuzzy rough sets]{\cite{radzikowska2002comparative}}
	\label{definition: ICFRS}
	Given $R\in \fpwset(X\times X)$ and $A\in\fpwset(X)$, the \emph{lower} and \emph{upper approximation} of $A$ w.r.t.\ $R$ are defined as:
	\begin{align}
		(\underline{\text{apr}}_{R} A)(x) & = \min\limits_{y\in X} \imp(R(x,y),A(y)), \label{lower approx}       \\
		(\overline{\text{apr}}_{R} A)(x)  & = \max\limits_{y\in X} \mathcal{C}(R(x,y),A(y)),\label{upper approx}
	\end{align}
	where $\imp$ is an implicator and $\mathcal{C}$ a conjunctor.
\end{definition}
The membership degree of an element \(x\) in the lower approximation of \(A\) can be interpreted (cf.\ \cite{theerens2023fqfrs}) as the truth value of the proposition ``All elements that are similar to \(x\) are in \(A\)''. An analogous interpretation exists for the upper approximation.
\begin{proposition}
	\label{prop: fuzzyroughMonotone}
	The lower and upper approximation satisfy relation monotonicity, i.e., for \(A\in\fpwset (X)\) we have
	\[(\forall R_1,R_2 \in \fpwset(X\times X))\left(R_1\subseteq R_2 \implies(\underline{apr}_{R_2}A) \subseteq (\underline{apr}_{R_1}A)\right).\]
\end{proposition}
For working with noisy datasets more robust fuzzy rough set models can be used \cite{theerens2022choquet,theerens2022fedcsis,THEERENS2024120385}. However, in general, for these extensions the lower approximation does not satisfy relation monotonicity.

Indiscernibility and similarity come naturally in decision systems. Decision systems are also a perfect framework for supervised learning.
\begin{definition}
	\label{def: decision system}
	An \emph{information system} $\left( X, \mathcal{A}\right)$, consists of a finite non-empty set \(X\) and a non-empty family of attributes $\mathcal{A}$, where each attribute \(a\in \mathcal{A}\) is a function $a:\ X \rightarrow V_a$, with $V_a$ the set of values the attribute $a$ can take.
	A \emph{decision system} is an information system $\left( X, \mathcal{A}\cup \{d\}\right)$, where \(d\notin \mathcal{A}\) is called the \emph{decision attribute} and each \(a\in\mathcal{A}\) is called a \emph{conditional attribute}.
\end{definition}
In supervised learning, the goal is to predict the decision attribute of a new instance by considering its conditional attributes and the information from the decision system. When the decision attribute consists of crisp classes, this is referred to as classification.

\subsection{Fuzzy-rough attribute measures}
Consider a family of similarity relations \(\{R_B: B\subseteq \mathcal{A}\}\) and a similarity relation \(R_d\) for the decision attribute. The authors in \cite{cornelis2010attribute} define the $B$-positive region $POS_B$ as the fuzzy set in $X$ defined as, for $y \in X$,
$$POS_{R_B}(y) = \left(\bigcup\limits_{x \in X}\underline{apr}_{R_B}(R_dx)\right)(y).$$
\begin{proposition}[\cite{cornelis2010attribute}]
	\label{prop: crispPositiveRegion}
	For \(y\in X\), if \(R_d\) is a crisp relation,
	\[POS_{R_B}(y) = \underline{apr}_{R_B}(R_dy)(y).\]
\end{proposition}
The value \(POS_{R_B}(y)\) can be interpreted as the degree to which similarity with respect to the conditional attributes \(B\) relates to similarity with respect to the decision attribute.
Consequently, the predictive ability of a subset \(B\) to predict the decision attribute \(d\), also called the degree of dependency of \(d\) on \(B\), is defined as:
$$\gamma_R(B) = \frac{|POS_{R_B}|}{|POS_{R_\mathcal A}|} = \frac{\sum\limits_{y \in X}POS_{R_B}(y)}{\sum\limits_{y \in X}POS_{R_\mathcal A}(y)}.$$
It is clear that this measure takes values between 0 and 1, and that \(\gamma_R(\mathcal{A}) = 1\).
A second measure considers the worst case scenario, i.e., to what extent there exists an element totally outside of the positive region:
$$\delta_R(B) = \frac{\min_{y \in X}POS_{R_B}(y)}{\min_{y \in X}POS_{R_\mathcal A}(y)}.$$

\subsection{Choquet integral}
The Choquet integral induces a large class of aggregation functions, namely the class of all comonotone linear aggregation functions \cite{beliakov2007aggregation}.
Since we will view the Choquet integral as an aggregation operator, we will restrict ourselves to measures (and Choquet integrals) on finite sets. For the general setting, we refer the reader to e.g.\ \cite{wang2010generalized}.
\begin{definition}
	Let \(X\) be a finite set. A function \(\mu:\mathcal{P}(X)\to[0,+\infty[\) is called a \emph{monotone measure} if:
	\[\mu(\emptyset)=0 \text{ and }\: (\forall A,B\in(\mathcal{P}(X))(A\subseteq B \implies \mu(A)\leq \mu(B))).\]
	A monotone measure is called \emph{additive} if \(\mu(A\cup B)=\mu(A)+\mu(B)\) when \(A\) and \(B\) are disjoint. If \(\mu(X)=1\), we call \(\mu\) normalized.
\end{definition}

\begin{definition}[\cite{wang2010generalized}]
	\label{defn: ChoquetIntegral}
	The \emph{Choquet integral} of \(f:X\to\mathbb{R}\) with respect to a monotone measure $\mu$ on \(X\) is defined as:
	\begin{equation*}
		\int f \diff \mu=\sum_{i=1}^n f(x^\ast_i)\cdot\left[\mu(A^\ast_i)-\mu(A^\ast_{i+1})\right],
	\end{equation*}
	where \((x^\ast_1,x^\ast_2,\dots,x^\ast_n)\) is a permutation of \(X=(x_1,x_2,\dots,x_n)\) such that
	\begin{equation*}
		f(x^\ast_1)\leq f(x^\ast_2) \leq\cdots\leq f(x^\ast_n),
	\end{equation*}
	\(A^\ast_i:=\{x^\ast_i,\dots,x^\ast_n\}\) and \(\mu(A^\ast_{n+1}):=0\).
\end{definition}
\begin{proposition}{\cite{beliakov2007aggregation}}
	\label{prop: choqAdditive}
	The Choquet integral with respect to an additive measure \(\mu\) is equal to
	\[\int f \diff \mu = \sum_{x\in X} f(x)\mu(\{x\}).\]
\end{proposition}
\begin{proposition}{\cite{wang2010generalized}}
	\label{prop: monotoneChoquet}
	Suppose \(\mu_1,\mu_2\) are two monotone measures with \(\mu_1 \leq \mu_2\), then for any \(f:X\to\mathbb{R}\) we have:
	\[\int f \diff \mu_1  \leq \int f \diff \mu_2.\]
\end{proposition}
\section{Fuzzy Rough Choquet Distances}
\label{sec: FRdistances}
In this section, we introduce fuzzy rough Choquet distances using the \(\gamma\) and \(\delta\) measures described in Section \ref{sec: fuzzy rough set theory}. Throughout this section, we will assume $\left( X, \mathcal{A}\cup \{d\}\right)$ is a decision system.
\subsection{Fuzzy rough Choquet p-distances}
\label{sec: general Choquet distance}
Inspired by Minkowski \(p\)-distances we define the Choquet \(p\)-distance as:
\begin{definition}[Choquet \(p\)-distance]
	Suppose \(\mu\) is a monotone measure on the set of conditional attributes \(\mathcal{A}\) and \(p\) is an integer.
	We define the Choquet \(p\)-distance \(d^\mu_p\) with respect to the monotone measure \(\mu\) as follows:
	\begin{equation}
		\label{eq: choquet distance}
		d^\mu_p(x,y) := \left(\int |a(x)-a(y)|^p \diff \mu(a)\right)^{\frac{1}{p}}.
	\end{equation}
\end{definition}
Note that when we use the counting measure \(\#(A)=|A|\), Eq.\ \eqref{eq: choquet distance} turns into the Minkowski p-distance (cf.\ Proposition \ref{prop: choqAdditive}):
\[d^\#_p(x,y) = \left(\sum_{a\in \mathcal{A}}|a(x)-a(y)|^p\right)^{\frac{1}{p}}.\]
This distance is in turn equal to the Manhattan (or Boscovich) distance if \(p=1\) and equal to the Euclidean distance if \(p=2\). In the limiting case of \(p\) reaching infinity we get the Chebyshev distance:
\[\lim_{p\to +\infty}d^\#_p(x,y) =\lim_{p\to +\infty} \left(\sum_{a\in \mathcal{A}}|a(x)-a(y)|^p\right)^{\frac{1}{p}}= \max_{a\in \mathcal{A}}|a(x)-a(y)|.\]
Similarly, if \(p\) goes to negative infinity, we get:
\[\lim_{p\to -\infty}d^\#_p(x,y) =\lim_{p\to -\infty} \left(\sum_{a\in \mathcal{A}}|a(x)-a(y)|^p\right)^{\frac{1}{p}}= \min_{a\in \mathcal{A}}|a(x)-a(y)|.\]
\vspace*{-5 mm}
\begin{remark}
	This definition varies slightly from previous definitions of Choquet distances (\cite{abril2012choquet,bolton2008discrete,torra2012comparison}), as it allows a more general set of distances beyond the ones that are based on the Euclidean distance (\(p=2\)). We made this choice to allow for greater flexibility, recognizing that in supervised learning scenarios, different distance metrics, especially the Manhattan distance (\(p=1\)), can be beneficial (cf.\ \cite{lenz2023unified}).
\end{remark}
The following example shows how allowing more general measures can be beneficial in a classification context.
\begin{example}
	\label{ex: 1}
	Consider the decision system illustrated in Table \ref{table:1}, presenting data on four patients, where the conditional attributes are fever, fatigue, and cough (each ranging from 0 to 1) and the decision attribute is the presence of flu.
	\begin{table}[H]
		\centering
		\def\arraystretch{1.1}%
		\setlength\tabcolsep{1.8 mm}
		\begin{tabular}{l |c |c| c ||c}
			        & \(a_1\) (fever) & \(a_2\) (fatigue) & \(a_3\) (cough) & \(d\) (flu) \\
			\hline
			\(x_1\) & 0               & 0.9               & 0.9             & 1           \\
			\(x_2\) & 0.9             & 0.95              & 0.95            & 1           \\
			\(x_3\) & 0               & 1                 & 0               & 0           \\
			\(x_4\) & 0.9             & 0                 & 0               & 0
		\end{tabular}
		\vspace*{2 mm}
		\caption{Decision system}
		\label{table:1}
	\end{table}
	\vspace*{-9 mm}
	Based on medical expertise, it is known that while most individuals with the flu exhibit several symptoms, not all symptoms need to be present. Additionally, it is known that not everyone with the flu will experience a fever.
	Therefore, an appropriate measure for constructing a Choquet distance could be:
	\begin{equation*}
		\begin{gathered}
			\mu(\{a_1\})=0.1, \mu(\{a_2\})=0.2 \text{ and } \mu(\{a_3\})=0.2, \\ \mu(\{a_1,a_2\})=0.2, \mu(\{a_1,a_3\})=0.2, \mu(\{a_2,a_3\})=0.8,\\
			\mu(\mathcal{A})=1, \mu(\emptyset)=0.
		\end{gathered}
	\end{equation*}
	Table \ref{table: 2} shows the distances calculated using the Choquet distance with respect to the measure \(\mu\), to the counting measure (i.e.\ the Manhattan distance) and with respect to the additive measure \(w\) defined by \(w(\{a_1\}) = 0.2\), \(w(\{a_2\}) = 0.4\), \(w(\{a_3\}) = 0.4\) (i.e.\ a distance based on a weighted average).
	\vspace*{-5 mm}
	\begin{table}[!htb]
		\def\arraystretch{1.1}%
		\setlength\tabcolsep{0.9 mm}
		\begin{subtable}{.33\linewidth}
		  \centering
		  \begin{tabular}{l |c |c| c| c}
			\(d^\mu_1\) & \(x_1\) & \(x_2\) & \(x_3\) & \(x_4\) \\
			\hline
			\(x_1\)     & 0.0     & 0.135   & 0.24    & 0.9     \\
			\(x_2\)     & 0.135   & 0.0     & 0.23    & 0.76    \\
			\(x_3\)     & 0.24    & 0.23    & 0.0     & 0.2     \\
			\(x_4\)     & 0.9     & 0.76    & 0.2     & 0.0     \\
		\end{tabular}
		\caption{Non-additive measure \(\mu\)}
		\end{subtable}%
		\begin{subtable}{.33\linewidth}
		  \centering
		  \begin{tabular}{l |c |c| c | c }
			\(d^\#_1\) & \(x_1\) & \(x_2\) & \(x_3\) & \(x_4\) \\
			\hline
			\(x_1\)    & 0.0     & 0.33    & 0.33    & 0.9     \\
			\(x_2\)    & 0.33    & 0.0     & 0.63    & 0.63    \\
			\(x_3\)    & 0.33    & 0.63    & 0.0     & 0.63    \\
			\(x_4\)    & 0.9     & 0.66    & 0.63    & 0.0
		\end{tabular}
		\caption{Counting measure \(\#\)}
		\end{subtable}%
		\begin{subtable}{.33\linewidth}
			\centering
			\begin{tabular}{l |c |c| c | c }
				\(d^w_1\) & \(x_1\) & \(x_2\) & \(x_3\) & \(x_4\) \\
				\hline
				\(x_1\)   & 0.0     & 0.22    & 0.4     & 0.9     \\
				\(x_2\)   & 0.22    & 0.0     & 0.58    & 0.76    \\
				\(x_3\)   & 0.4     & 0.58    & 0.0     & 0.58    \\
				\(x_4\)   & 0.9     & 0.76    & 0.58    & 0.0
			\end{tabular}
			\caption{Additive measure \(w\)}
		  \end{subtable} 
		  \vspace*{2 mm}
		\caption{Choquet distances with respect to several measures}
		\label{table: 2}
	\end{table}
	\vspace*{-8 mm}
				
	According to both \(d^\#_1\) and \(d^w_1\), \(x_1\) is considered the nearest neighbour of \(x_3\). However, when considering their similarity in terms of flu symptoms, \(x_3\) appears to be closer to \(x_4\) than \(x_1\). This is because the fever attribute plays a less prominent role in determining similarity in this context. In contrast, the Choquet distance correctly identifies \(x_4\) as the nearest neighbour to \(x_3\). In this example, the Choquet distance consistently ensures that the nearest neighbour belongs to the same decision class, whereas the Manhattan distance performs poorly, and the weighted distance shows marginal improvement.
\end{example}
The challenge now is to select a suitable/adequate measure \(\mu\) for use in Eq.\ \eqref{eq: choquet distance} to define a concrete distance. In the absence of expert knowledge such as the medical information used in Example \ref{ex: 1}, one solution is to use the \(\gamma\) and \(\delta\) measures from rough set theory. These measures indicate the importance of attribute subsets in predicting the decision attribute, making them suitable for supervised learning. In order for the Choquet integral to be defined unambiguously, it is necessary for these measures to be monotone. Therefore, we will investigate under which conditions these fuzzy rough attribute measures are monotone.
\begin{definition}
	A family of similarity relations \(\{R_B: B\subseteq \mathcal{A}\}\) is called monotone if
	\[(\forall A,B \in \pwset(\mathcal{A}))(A\subseteq B \implies R_B \subseteq R_A).\]
\end{definition}
\begin{example}
	We can extend every family of similarity relations \(\{R_{a} : a \in \mathcal{A}\}\) to a monotone family \(\{R_B: B\subseteq \mathcal{A}\}\) as follows:
	\[R_B(x,y) := \mathcal{T}_{a\in B} (R_a(x,y)).\]
	Or more general, suppose \(Ag\) is an aggregation operator:
	\[R_B(x,y) := Ag_{a\in \mathcal{A}}(\phi^{x,y}_B(a)),\]
	where for each \(a\in\mathcal{A}\),
	\[\phi^{x,y}_B(a):= \begin{cases}
			R_a(x,y), & \text{if $a\in\mathcal{A}$}, \\
			1         & \text{otherwise}.
		\end{cases}\]
\end{example}
\begin{proposition}
	The degree of dependency \(\gamma_R\) and \(\delta_R\) are monotone measures if the family of similarity relations \(\{R_B: B\subseteq \mathcal{A}\}\) is monotone.
\end{proposition}
\begin{proof}
	Follows from the monotonicity of the lower approximation with respect to the relation (Proposition \ref{prop: fuzzyroughMonotone}):
	\begin{align*}
		A\subseteq B & \implies R_B \subseteq R_A \implies \underline{apr}_{R_A}(R_dx) \subseteq\underline{apr}_{R_B}(R_dx)                                                                                  \\
		             & \implies (\forall y \in X) \left( \left(\bigcup\limits_{x \in X}\underline{apr}_{R_A}(R_dx)\right)(y) \leq \left(\bigcup\limits_{x \in X}\underline{apr}_{R_B}(R_dx)\right)(y)\right) \\
		             & \implies \gamma_R(A)\leq \gamma_R(B) \text{ and } \delta_R(A)\leq \delta_R(B)
	\end{align*}
\end{proof}
If the attribute dependency measures lack monotonicity due to the absence of monotonicity in the family of relations or the use of a robust fuzzy rough set model (resulting in some cases to the loss of monotonicity in the lower approximation), we can resolve this issue by either monotonizing the measure or by applying monotonization to the family of similarity relations. It is worth mentioning that the second approach does not work in the case we replace the classical fuzzy rough set with robust fuzzy rough set models that lack monotonicity (cf.\ \cite{theerens2023fqfrs}).
\begin{definition}
	We define the monotonization \(\{R^m_B: B\subseteq \mathcal{A}\}\) of a family of similarity relations \(\{R_B: B\subseteq \mathcal{A}\}\) as:
	\[R^m_A(x,y) := \min_{S\subseteq A} R_S(x,y).\]
	Analogously, we define the monotonization \(\mu^m\) of a set function \(\mu:\pwset(\mathcal{A})\to [0, +\infty[\) as:
	\[\mu^m(B) := \frac{\max_{S\subseteq B} \mu(S)}{\max_{A\subseteq \mathcal{A}} \mu(A)}.\]
\end{definition}
\begin{remark}
	More generally we could define these monotonization using t-norms and t-conorms.
\end{remark}
Note that the monotonization \(\mu^m\) of a set function \(\mu:\pwset(\mathcal{A})\to [0, +\infty[\) not only ensures monotonicity but also normalizes it (\(\mu^m(\mathcal{A})=1\)). Consequently, \(\mu^m\) is always a monotone measure.

In general, we have these inequalities between the different measures and the monotonizations of them:
\begin{proposition}
	If \(POS_{R_{\mathcal{A}}}(y)= 1\) for all \(y\in X\), then it holds that:
	\[(\delta_R)^m \leq \delta_{R^m},(\gamma_R)^m \leq \gamma_{R^m}\text{ and }\;\delta \leq \gamma.\]
\end{proposition}
\begin{proof}
	Since \(R_1 \subseteq R_2 \implies POS_{R_1}\supseteq POS_{R_2}\) we have that:
	\begin{align*}
		(\delta_R)^m (B) & = \frac{\max\limits_{S\subseteq B} \delta_R(S)}{\max\limits_{S\subseteq \mathcal{A}} \delta_R(S)}= \frac{\max\limits_{S\subseteq B}  \frac{\min\limits_{y \in X}POS_{R_S}(y)}{\min\limits_{y \in X}POS_{R_{\mathcal{A}}}(y)}               }{\max\limits_{S\subseteq \mathcal{A}}  \frac{\min\limits_{y \in X}POS_{R_S}(y)}{\min\limits_{y \in X}POS_{R_{\mathcal{A}}}(y)}               } \\
		                 & = \frac{\max\limits_{S\subseteq B} \min\limits_{y \in X}POS_{R_S}(y)}{\max\limits_{S\subseteq \mathcal{A}} \min\limits_{y \in X}POS_{R_S}(y)}
		\leq \frac{\max\limits_{S\subseteq B} \min\limits_{y \in X}POS_{R^m_B}(y)}{\max\limits_{S\subseteq \mathcal{A}} \min\limits_{y \in X}POS_{R_S}(y)}                                                                                                                                                                                                                                                            \\
		                 & =\frac{ \min\limits_{y \in X}POS_{R^m_B}(y)}{\max\limits_{S\subseteq \mathcal{A}} \min\limits_{y \in X}POS_{R_S}(y)}\leq\frac{\min\limits_{y \in X}POS_{R^m_B}(y)}{\min\limits_{y \in X}POS_{R_{\mathcal{A}}}(y)},
	\end{align*}
	as well as
	\[1=POS_{R_{\mathcal{A}}}(y)\leq POS_{R^m_{\mathcal{A}}}(y)\leq 1\]
	Combining these two gives us the desired result:
	\begin{align*}
		(\delta_R)^m (B) & \leq \frac{\min\limits_{y \in X}POS_{R^m_B}(y)}{\min\limits_{y \in X}POS_{R_{\mathcal{A}}}(y)} = \frac{\min\limits_{y \in X}POS_{R^m_B}(y)}{\min\limits_{y \in X}POS_{R^m_{\mathcal{A}}}(y)} =\delta_{R^m} (B).
	\end{align*}
	The proof goes analogously for \((\gamma_R)^m \leq \gamma_{R^m}\). The proof for \(\delta \leq \gamma\) is straightforward.
\end{proof}
These inequalities extend to the distances as well, as Choquet integrals exhibit monotonicity with respect to the measure (cf.\ Proposition \ref{prop: monotoneChoquet}). For instance, suppose we have two measures \(\mu_1\) and \(\mu_2\) with \(\mu_1 \leq \mu_2\), then the distance between two points \(x,y \in X\) based on \(\mu_1\) will always be smaller than the distance between these points determined by \(\mu_2\):
\begin{align*}
	d^{\mu_1}_p(x,y) & = \left(\int |a(x)-a(y)|^p \diff \mu_1(a)\right)^{\frac{1}{p}} \\ &\leq \left(\int |a(x)-a(y)|^p \diff \mu_2(a)\right)^{\frac{1}{p}} = d^{\mu_2}_p(x,y).
\end{align*}
\subsection{Choquet distances for classification}
\label{sec: Choquet distance classification}
In the case of classification, we have that $R_d$ is a crisp equivalence relation. In this case (using Proposition \ref{prop: crispPositiveRegion}) we can simplify the positive region as follows:
\begin{align*}
	POS_{R_B}(y)
	 & =\underline{apr}_{R_B}(R_dy)(y) = \min\limits_{z \in X}I(R_B(z,y),R_d(z,y))                                      \\
	 & = \min\left(\min\limits_{z \in R_dy}I(R_B(z,y),R_d(z,y)), \min\limits_{z \notin R_dy}I(R_B(z,y),R_d(z,y))\right) \\
	 & = \min\left(\min\limits_{z \in R_dy}I(R_B(z,y),1), \min\limits_{z \notin R_dy}I(R_B(z,y),0)\right)               \\
	 & = \min\limits_{z \not \in R_dy}I(R_B(z,y),0)  = \min\limits_{z \notin R_dy}N_I(R_B(z,y))                         \\
	 & =\min\limits_{z \notin R_dy}d^I_B(z,y),
\end{align*}
where \(d^I_B(x,y):= N_I(R_B(x,y))\) for all \(x,y \in X\). Note that since \(R_B\) represents a similarity measure, \(d^I_B\) can be interpreted as a distance measure. This leads us to generalize our \(\gamma\) and \(\delta\) measure as follows:
\begin{equation}
	\label{eq: gammadelta}
	\gamma_d(B) =  \frac{\sum\limits_{y \in X}\min\limits_{z \notin R_dy}d_B(z,y)}{\sum\limits_{y \in X}\min\limits_{z \notin R_dy}d_{\mathcal{A}}(z,y)} \;\text{ and }\;
	\delta_d(B) =\frac{\min\limits_{y \in X}\min\limits_{z \notin R_dy}d_B(z,y)}{\min\limits_{y \in X}\min\limits_{z \notin R_dy}d_{\mathcal{A}}(z,y)}.
\end{equation}
\begin{definition}
	A family of distances \(\left\{d_B:X\times X \to [0,+\infty[ \,:\, B\subseteq \mathcal{A}\right\}\) is called monotone if
	\[(\forall A,B \in \pwset(\mathcal{A}))(A\subseteq B \implies d_A \leq d_B).\]
\end{definition}
\begin{proposition}
	The degree of dependency \(\gamma_d\) and \(\delta_d\) are monotone measures if the family of distances \(\{d_B: B\subseteq \mathcal{A}\}\) is monotone.
\end{proposition}
The interpretation of Eq.\ \eqref{eq: gammadelta} is that the dependency of a subset \(B\) towards the decision attribute can be interpreted as the normalized average (or minimum in the case of \(\delta\)) of the distances to the closest out-of-class instance.

This more general definition makes it easier to construct measures for classification:
\begin{example}
	Using the Manhattan distance \(d_B(x,y) = \sum\limits_{a\in B}|a(z)-a(y)|\), we get
	\begin{equation}
		\label{eq: gammaClass}
		\gamma_d(B) =  \frac{\sum\limits_{y \in X}\min\limits_{z \notin R_dy}\left(\sum\limits_{a\in B}|a(z)-a(y)|\right)}{\sum\limits_{y \in X}\min\limits_{z \notin R_dy}\left(\sum\limits_{a\in \mathcal{A}}|a(z)-a(y)|\right)}.
	\end{equation}
	Of course the Manhattan distance could be replaced with any other distance. Applying Equation \eqref{eq: gammaClass} to the decision system of Example \ref{ex: 1}, we get
	\begin{equation}
		\label{eq: gammaconcrete}
		\begin{gathered}
			\gamma_d(\{a_1\})=0.0, \gamma_d(\{a_2\})=0.19 \text{ and } \gamma_d(\{a_3\})=0.63, \\ \gamma_d(\{a_1,a_2\})=0.36, \gamma_d(\{a_1,a_3\})=0.64, \gamma_d(\{a_2,a_3\})=0.83,\\
			\gamma_d(\mathcal{A}))=1, \gamma_d(\emptyset)=0.
		\end{gathered}
	\end{equation}
	The Choquet distances calculated using the \(\gamma\) measure from Eq. \eqref{eq: gammaconcrete} are displayed in Table \ref{table:5}.
	\vspace*{-5 mm}
	\begin{table}[H]
		\centering
		\def\arraystretch{1.1}%
		\setlength\tabcolsep{0.9 mm}
		\begin{tabular}{l |c |c| c | c }
			\(d^\gamma_1\) & \(x_1\) & \(x_2\) & \(x_3\) & \(x_4\) \\
			\hline
			\(x_1\)   & 0.0 & 0.05 & 0.59 & 0.9 \\
			\(x_2\)   & 0.05 & 0.0 & 0.62 & 0.79 \\
			\(x_3\)   & 0.59 & 0.62 & 0.0 & 0.34 \\
			\(x_4\)   & 0.9 & 0.79 & 0.34 & 0.0
		\end{tabular}
		\vspace*{2 mm}
		\caption{Fuzzy rough Choquet distance using \(\gamma\)}
		\label{table:5}
	\end{table}
	\vspace*{-11 mm}
	This distance outperforms the Choquet distance with respect to \(\mu\) used in Example \ref{ex: 1}, as it brings instances of the same class closer together while pushing instances of different classes farther apart.
\end{example}
We conclude this section by outlining how Choquet distances can be employed for classification through a step-by-step procedure.
\begin{enumerate}
	\item Calculate the \(\gamma\) (or \(\delta\)) measure based on a selected distance metric (e.g., the Manhattan distance, see Eq. \eqref{eq: gammaClass}).
	\item Determine the Choquet \(p\)-distance (for a chosen \(p\)) based on the calculated measure.
	\item Classify new instances using a distance-based classification algorithm, such as KNN, and use the Choquet distance from the previous step to calculate the distances.
\end{enumerate}
\section{Conclusion and future research}
\label{sec: conclusion}
In summary, this paper introduces concrete monotone measures designed for application with the Choquet integral, aiming to define distances suitable for classification tasks. We have studied two different measures, two ways to monotonize them, and showed some inequalities between them. As these measures have the interpretation of attribute importance towards the decision attribute, we may claim that these distances are intuitive and promising.

However, there is room for future exploration. Experimental evaluations, such as applying these distances in classification tasks on benchmark datasets using KNN, are needed. Additionally, further investigation into alternative measures that evaluate attribute importance in subsets of conditional attributes, such as ones based on fuzzy quantifiers or information theory, could lead to improved Choquet distances. For greater expressiveness, one may explore the use of the more general Choquet-Mahalanobis operator introduced in \cite{torra2012comparison} to define a distance for supervised learning. Finally, an approach inspired by distance metric learning \cite{suarez2021tutorial}, where the measure is learned through optimizing a loss function could be interesting.
%
%
%
\bibliographystyle{splncs04}
\bibliography{/Users/adnantheerens/Desktop/phd/Articles/bibfiles/bibfile}

\end{document}